\documentclass{article}

\usepackage{arxiv}
\usepackage{mathpazo}
\usepackage[utf8]{inputenc}
\usepackage[T1]{fontenc}

\usepackage{amsmath}
\usepackage{amsfonts}

\usepackage{array}
\usepackage{booktabs}
\usepackage{multirow}
\usepackage{float}

\usepackage{graphicx}
\usepackage{xcolor}

\usepackage{nicefrac}
\usepackage{microtype}

\usepackage{natbib}
\usepackage{doi}
\usepackage{url}
\usepackage{hyperref}

\title{Rethinking Evaluation for Video Capsule Endoscopy: Validation Protocol, Source-Video Diversity, and Cross-Dataset Shift}

\author{
\href{https://orcid.org/0009-0008-9629-0650}
{\includegraphics[scale=0.06]{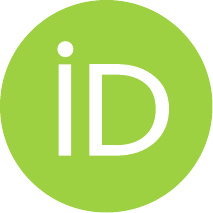}\hspace{1mm}Smriti Regmi} \\
Department of Computer Science\\
Biomedical Perception \& Intelligence Lab\\
University of South Dakota\\
Vermillion, South Dakota, USA\\
\texttt{smriti.regmi@coyotes.usd.edu}
\And
\href{https://orcid.org/0000-0002-8078-6730}
{\includegraphics[scale=0.06]{orcid.pdf}\hspace{1mm}Debesh Jha} \\
Department of Computer Science\\
Biomedical Perception \& Intelligence Lab\\
University of South Dakota\\
Vermillion, South Dakota, USA\\
\texttt{debesh.jha@usd.edu}
}

\renewcommand{\shorttitle}{Rethinking Evaluation for Video Capsule Endoscopy}

\hypersetup{
pdftitle={Rethinking Evaluation for Video Capsule Endoscopy: Validation Protocol, Source-Video Diversity, and Cross-Dataset Shift},
pdfsubject={Computer Vision, Medical Imaging, Video Capsule Endoscopy},
pdfauthor={Smriti Regmi, Debesh Jha},
pdfkeywords={video capsule endoscopy, rare disease classification},
}

\begin{document}
\maketitle

\begin{abstract}
Automated video capsule endoscopy (VCE) classification remains challenging due to severe class imbalance, limited source-video diversity, and variations in image appearance across datasets. Although deep learning models have demonstrated promising performance on established benchmarks, their reliability under different validation protocols and external dataset shift remains insufficiently understood. In this work, we introduce DMSNet, a deformable adaptive multi-scale edge-guided attention network integrating adaptive multi-scale feature fusion, deformable lesion refinement, edge-guided attention, and sparse feature selection. We benchmark DMSNet against 18 representative convolutional, transformer-based, and pretrained vision architectures on Kvasir-Capsule under random frame-level and source-video-disjoint validation protocols. We further identify source-video overlap between the original training and test splits and evaluate model performance on both the complete test set and a source-video-disjoint  subset. Under random frame-level validation, DMSNet achieves an MCC of 0.442 ± 0.020 on complete test set and 0.456 ± 0.037 on the source-video-disjoint  subset. On the external CV2024 dataset, ConvNeXt-Tiny achieves the highest MCC of 0.207, while performance degrades substantially across architectures relative to in-domain evaluation. We further investigate source-video diversity using a per-class cap of 200 training frames. With the total training-set size held fixed, increasing the maximum source-video coverage from four to eight improves MCC for 17 of 19 evaluated architectures.Overall, these findings demonstrate that model rankings and classification performance depend substantially on evaluation protocol, training-source diversity, and cross-dataset shift, emphasizing the need for source-aware validation and rigorous external evaluation of VCE classification systems.
\end{abstract}

\section{Introduction}
\label{sec:intro}
Video capsule endoscopy (VCE) is a non-invasive imaging technique that enables visualization of the gastrointestinal (GI) tract, particularly regions such as the small intestine that are difficult to examine using conventional endoscopy. During an examination, a swallowable capsule equipped with a miniature camera captures tens of thousands of images, allowing clinicians to identify abnormalities such as ulcers, erosions, and angiectasia. However, reviewing such large volumes of image data is time-consuming and places a substantial workload on
gastroenterologists. The large number of frames, together with
subtle pathological findings and variations in image quality,
also makes reliable interpretation challenging. Consequently,
computer-aided diagnosis (CAD) systems have received increasing
attention as a means of supporting efficient and consistent
VCE image analysis. Recent advances in convolutional neural networks (CNNs), vision transformers (ViTs), and large-scale pretrained vision models have substantially improved classification performance. CNN-based architectures~\cite{he2016deep,howard2019searching,huang2017densely,radosavovic2020designing,tan2021efficientnetv2,tan2019efficientnet,woo2023convnext,liu2022convnet} effectively capture local spatial features, while vision transformer-based models~\cite{dosovitskiy2020image,tu2022maxvit, touvron2021going, touvron2021training,liu2021swin,liu2022swin} offer stronger global context modeling through self-attention. More recently, self-supervised foundation models~\cite{simeoni2025dinov3,fang2024eva,wang2023foundation} pretrained on large-scale natural images or on endoscopic data have demonstrated promising transfer capabilities. However, strong performance under conventional frame-level validation does not necessarily indicate reliable generalization in video capsule endoscopy. Frames originating from the same source video can exhibit substantial visual redundancy, while individual gastrointestinal findings may be represented by only a limited number of independent source videos. Consequently, the number of annotated frames alone may not accurately reflect the diversity available for learning generalizable representations. Random frame-level partitioning can place highly correlated frames from the same source video in both the training and validation sets.  Moreover,  strong performance on an in-domain benchmark may not translate to the data collected under different acquisition protocols. These considerations motivate a broader investigation of how model architecture, validation strategy, and source-video diversity influence measured generalization in VCE classification. In this work, we investigate VCE classification from both architectural and generalization perspectives. We introduce DMSNet, a deformable multi-scale network built on an EfficientNetV2-S~\cite{tan2021efficientnetv2} backbone. The proposed architecture integrates adaptive multi-scale feature fusion, deformable lesion refinement, edge-guided attention, and sparse feature selection to capture complementary spatial and semantic representations. Unlike conventional fixed feature aggregation, the adaptive fusion module learns input-dependent weights for combining features extracted at different backbone stages. We evaluate DMSNet alongside 18 representative CNN, transformer, and pretrained vision architectures using random frame-level and source-video-disjoint validation strategies on Kvasir-Capsule. In addition to comparing validation protocols, we examine performance on both the original test set and a source-video-disjoint test subset obtained by excluding frames whose source-video identifiers overlap with those in the training fold. We further investigate source-video diversity by imposing a per-class cap of 200 training frames and varying the maximum number of source videos from one to eight. Although data-limited classes may not always reach the frame cap, the four- and eight-video settings contain the same total number of training frames, enabling the cleanest comparison of broader source-video coverage without increasing the overall training-set size. Finally, we assess cross-dataset generalization on the eight categories shared between Kvasir-Capsule and CV2024.

The main contributions of this work are as follows:
\begin{enumerate}
\item We conduct a systematic evaluation of DMSNet and 18 representative CNN, transformer, and pretrained vision architectures under random frame-level and source-video-disjoint validation. We further compare their performance on the original Kvasir-Capsule test set and a source-video-disjoint test subset, demonstrating how evaluation protocols influence model rankings and class-wise performance
\item We study source-video diversity across 19 architectures and show that, with the total training-set size fixed at 2,010 frames, increasing the maximum source-video coverage from four to eight improves mean MCC for 17 of 19 models. A paired one-sided Wilcoxon signed-rank test confirms a significant positive shift across architectures.
\item We evaluate Kvasir-Capsule-trained models on the eight categories shared with CV2024 and observe substantial cross-dataset performance degradation, showing that strong in-domain performance does not consistently translate to stronger external performance.
\item We introduce DMSNet, a competitive deformable multi-scale edge-guided attention network combining multi-scale feature fusion, deformable refinement, edge-guided attention, and sparse feature selection.
\end{enumerate}

\section{Related Work}
\label{sec:related_work}
\subsection{Deep Learning for Capsule Endoscopy}
Deep learning has been widely adopted for automated classification of gastrointestinal findings in video capsule endoscopy (VCE). Kvasir-Capsule~\cite{smedsrud2021kvasir} is a common benchmark, with subsequent work exploring focal modulation~\cite{srivastava2022video}, explainable AI~\cite{varam2023wireless}, binary
classification~\cite{oukdach2024vitca}, and semi-supervised learning~\cite{muruganantham2025uncertainty}. CNNs, vision transformers
such as ViT~\cite{dosovitskiy2020image} and
DeiT~\cite{touvron2021training}, and pretrained models including
DINOv3~\cite{simeoni2025dinov3}, EVA-02~\cite{fang2024eva}, and
Endo-FM~\cite{wang2023foundation} have all been investigated for visual
representation learning. Despite these advances,
VCE classification remains challenging because of severe class imbalance and limited independent source-video diversity.
\subsection{Imbalanced and Multi-Scale Feature Learning}
Class imbalance is a persistent challenge in medical imaging. Common strategies include over-sampling
\cite{krawczyk2012oversampling,asaduzzaman2024machine,he2008adasyn,chawla2002smote},
focal loss~\cite{lin2017focal}, class-balanced
loss~\cite{cui2019class}, LDAM~\cite{cao2019learning}, logit
adjustment~\cite{zhang2024gradient}, and classifier
re-training~\cite{kang2019decoupling,lu2024rethinking}. In medical video datasets, however, large frame counts do not necessarily imply high
independent sample diversity because many visually similar frames may originate from the same recording.
Complementary architectural work has investigated multi-scale and attention-based feature learning using spatial pyramids
\cite{he2014spatial,lin2017feature,zhao2017pyramid}, channel and spatial attention~\cite{hu2018squeeze,woo2018cbam}, and deformable convolutions~\cite{dai2017deformable,zhu2019deformable}. DMSNet builds on these ideas by combining multi-scale adaptive feature fusion, deformable refinement,
edge-guided attention, and sparse feature selection.
\subsection{Evaluation Protocol and Model Selection}
The importance of grouping correlated medical images during dataset partitioning is well established.~\cite{tampu2022inflation} showed that per-image splitting of correlated OCT images can inflate estimated performance relative to subject-wise partitioning, while~\cite{bussola2021ai} reported similar leakage effects in digital pathology. In VCE, we therefore use source-video grouping as the available unit of independence. Our contribution is not to rediscover
image-level leakage, but to quantify how validation strategy changes relative model rankings and conclusions across VCE architectures. Concurrent work~\cite{hanson2026benchmarking} studies cross-target model-selection instability by evaluating 11 pretrained backbones across Kvasir-Capsule, CV2024, and Galar. In our study, we compare random frame-level and source-video-disjoint validation
within Kvasir-Capsule, control source-video diversity under a capped
frame budget, and evaluate the resulting models under external shift on CV2024.

\section{Methodology}
\label{sec:methodology}

\subsection{Backbone and Adaptive Multi-Scale Feature Fusion}
We adopt EfficientNetV2-S as the feature
extraction backbone. Given an input image
$\mathbf{x}\in\mathbb{R}^{3\times224\times224}$, features are extracted
from four stages of increasing depth:
\begin{equation}
\left\{
\mathbf{F}^{(1)},\mathbf{F}^{(2)},
\mathbf{F}^{(3)},\mathbf{F}^{(4)}
\right\}
=
\mathcal{B}(\mathbf{x}),
\label{eq:backbone}
\end{equation}

where $\mathcal{B}(\cdot)$ denotes the EfficientNetV2-S backbone.
Shallow features retain fine-grained spatial and boundary information,
whereas deeper features encode higher-level semantic representations.
\begin{figure*}[htbp]
\centering
\includegraphics[width=\linewidth]{
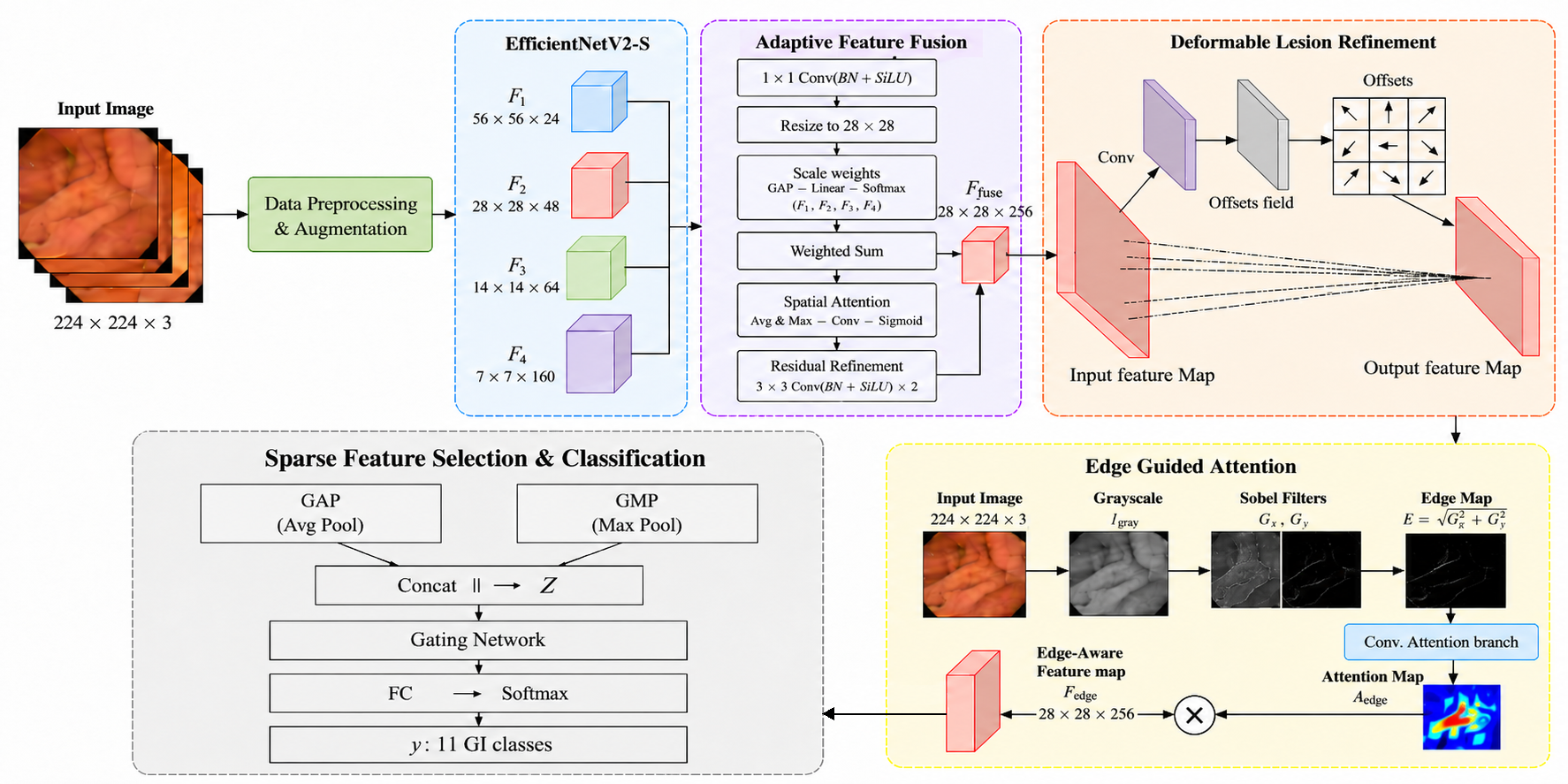}
   \caption{Overall architecture of the proposed DMSNet framework.}
\label{fig:architecture}
\end{figure*}
Figure~\ref{fig:architecture} illustrates the overall architecture of the proposed framework, consisting of six major components: EfficientNetV2-S backbone, Adaptive Multi-Scale Feature Fusion (AMSFF), Deformable Lesion Refinement (DLR), Edge-Guided Attention (EGA), Sparse Feature Selection (SFS), and the final classification head.

To integrate information from different feature extraction stages, we implement an adaptive fusion mechanism that dynamically determines the contribution of each scale. First, each backbone feature map is projected to a channel dimension of 256 using a $1\times1$ convolution-BatchNorm-SiLU block. The projected features are then aligned through bilinear interpolation to the spatial resolution of the second selected backbone feature map:
\begin{equation}
\mathbf{P}^{(i)}
=
\mathrm{Resize}_{H_2,W_2}
\left(
\mathrm{ConvBNAct}_{1\times1}
(\mathbf{F}^{(i)})
\right),
\quad i=1,\ldots,4,
\label{eq:adaptive_projection}
\end{equation}

where $H_2$ and $W_2$ denote the spatial dimensions of
$\mathbf{F}^{(2)}$.

Rather than concatenating the aligned features, the proposed
module learns an input-dependent importance weight for each scale.
A global average pooling operation generates a channel descriptor
for each aligned feature map, which is subsequently passed through
a scale-specific linear layer:

\begin{equation}
s_i
=
\mathbf{w}_i^{\top}
\mathrm{GAP}(\mathbf{P}^{(i)})
+b_i,
\label{eq:scale_score}
\end{equation}

where $s_i$ represents the importance score of the $i$-th scale.
The scores are normalized using a softmax function:

\begin{equation}
\beta_i
=
\frac{\exp(s_i)}
{\sum_{j=1}^{4}\exp(s_j)},
\label{eq:scale_weights}
\end{equation}

where $\beta_i$ denotes the normalized contribution of the
$i$-th feature scale. The aligned representations are then
aggregated through weighted summation:

\begin{equation}
\mathbf{F}_{w}
=
\sum_{i=1}^{4}
\beta_i\mathbf{P}^{(i)}.
\label{eq:weighted_fusion}
\end{equation}

To further refine the spatial information within the fused
representation, a spatial attention mechanism is applied.
Channel-wise average and maximum pooling are performed on
$\mathbf{F}_{w}$, and the resulting spatial descriptors are
concatenated and processed using a $7\times7$ convolution
followed by a sigmoid activation:

\begin{equation}
\mathbf{A}_{s}
=
\sigma
\left(
\mathrm{Conv}_{7\times7}
\left(
[
\mathrm{Avg}_{c}(\mathbf{F}_{w})
\Vert
\mathrm{Max}_{c}(\mathbf{F}_{w})
]
\right)
\right),
\label{eq:spatial_attention}
\end{equation}

where $\mathrm{Avg}_{c}$ and $\mathrm{Max}_{c}$ denote
channel-wise average and maximum pooling, respectively.
The spatially attended representation is computed as

\begin{equation}
\mathbf{F}_{a}
=
\mathbf{F}_{w}\odot\mathbf{A}_{s}.
\label{eq:spatial_refinement}
\end{equation}

Finally, the attended features are processed using two
successive $3\times3$ convolutional blocks. The first block
includes BatchNorm and SiLU activation, while the second
includes BatchNorm without an activation. A residual
connection from the original weighted fused representation
is used to preserve the aggregated multi-scale information:

\begin{equation}
\mathbf{F}_{fus}
=
\mathrm{SiLU}
\left(
\mathbf{F}_{w}
+
\mathcal{R}(\mathbf{F}_{a})
\right),
\label{eq:adaptive_fusion}
\end{equation}

where $\mathcal{R}(\cdot)$ denotes the two-block convolutional
refinement operation. The resulting feature representation
contains 256 channels at the spatial resolution of the
second selected backbone feature map.

\subsection{Deformable Lesion Refinement}

Gastrointestinal abnormalities often exhibit irregular shapes and boundaries.
To adapt the receptive field to such spatial variations, we apply deformable
convolution~\cite{dai2017deformable} to the fused representation:

\begin{equation}
\mathbf{F}_{def}(\mathbf{p})
=
\sum_{k=1}^{K}
w_k
\mathbf{F}_{fus}
\left(
\mathbf{p}+\mathbf{p}_k+\Delta\mathbf{p}_k
\right),
\label{eq:deform}
\end{equation}

where $\mathbf{p}_k$ denotes the regular sampling location and
$\Delta\mathbf{p}_k$ is a learned offset. The deformable representation
is combined with the original fused features through a residual connection,

\begin{equation}
\mathbf{F}_{r}
=
\mathbf{F}_{fus}+\mathbf{F}_{def}.
\label{eq:residual}
\end{equation}

\subsection{Edge-Guided Attention}

To incorporate explicit boundary information, the input image is converted
to grayscale and processed using Sobel operators. The resulting edge magnitude
$E=\sqrt{G_x^2+G_y^2}$ is passed through a lightweight convolutional branch
to generate a spatial attention map:

\begin{equation}
A=\sigma(\psi(E)),
\qquad
\mathbf{F}_{att}
=
\mathbf{F}_{r}
+
\mathbf{F}_{r}\odot A,
\label{eq:edgeattention}
\end{equation}

where $\psi(\cdot)$ denotes the convolutional attention branch,
$\sigma(\cdot)$ is the sigmoid function, and $\odot$ denotes element-wise
multiplication. The residual formulation emphasizes boundary-aware regions
while preserving the underlying semantic representation.

\subsection{Sparse Feature Selection and Classification}
The feature representation is summarized by combining Global Average Pooling (GAP) and Global Max Pooling (GMP):

\begin{equation}
\mathbf{z}
=
[
\mathrm{GAP}(\mathbf{F}_{att})
\Vert
\mathrm{GMP}(\mathbf{F}_{att})
].
\label{eq:pool}
\end{equation}

A two-layer gating network then produces a feature-selection mask,

\begin{equation}
\mathbf{g}
=
\sigma
\left(
W_2\delta(W_1\mathbf{z}+b_1)+b_2
\right),
\qquad
\mathbf{z}_{sel}
=
\mathbf{g}\odot\mathbf{z},
\label{eq:selection}
\end{equation}

where $\delta(\cdot)$ denotes the SiLU activation. The gated representation
is passed to a linear classifier to produce logits for the $K=11$ classes:

\begin{equation}
\mathbf{o}
=
W_c\mathbf{z}_{sel}+b_c.
\label{eq:logits}
\end{equation}

The network is optimized using focal loss together with mean gate-activation sparsity penalty,
\begin{equation}
\mathcal{L}
=
\mathcal{L}_{\mathrm{Focal}}
+
\lambda \mathcal{L}_{\mathrm{Sparse}},
\qquad
\mathcal{L}_{\mathrm{Sparse}}
=
\frac{1}{d}\sum_{j=1}^{d} g_j,
\end{equation}
where $g_j$ denotes the $j$-th sigmoid gate activation and $d$ is the
number of gated feature dimensions. We set $\lambda=10^{-3}$ in all
DMSNet experiments. The sparsity term penalizes the mean gate activation, encouraging suppression of less relevant feature dimensions.
\section{Experiments}
We evaluate DMSNet and 18 representative CNN, transformer, and pretrained vision architectures under three experimental settings: (1) in-domain evaluation on Kvasir-Capsule using random frame-level and source-video-disjoint validation, with performance assessed on both the original test set and a source-video-disjoint test subset; (2) a controlled source-video diversity study with a per-class cap of 200 training frames; and (3) cross-dataset evaluation on CV2024 using eight shared categories.

\subsection{Datasets}
\subsubsection{Kvasir-Capsule}
Kvasir-Capsule contains 47,238 labeled frames from 14 original categories. We construct an 11-class task by retaining \textit{Blood-fresh} as
\textit{Blood} and excluding \textit{Blood-hematin},
\textit{Ampulla of Vater}, and \textit{Polyp}. The resulting classes are Normal, Pylorus, Ileo-cecal Valve, Angiectasia, Blood, Erosion,
Erythematous, Foreign Bodies, Lymphangiectasia, Reduced Mucosal View, and Ulcer. Source-video identifiers are extracted from frame filenames and used to construct group-aware data partitions.In this study, source-video identifiers correspond to individual VCE recordings rather than verified patient identities; therefore, source-video-disjoint partitioning does not necessarily imply patient-level independence.
 \subsubsection{CV2024}
For external evaluation, we use the independent test set of CV2024~\cite{handa2024capsule}, collected at the All India Institute of Medical Sciences (AIIMS), India. Although the CV2024 training set includes Kvasir-Capsule data, its test set comprises independently collected AIIMS data. Evaluation is restricted to the eight semantically corresponding classes shared between the two datasets, with the complete label mapping provided in the supplementary material.
\subsection{Models and Training Setup}
\label{sec:models and training setup}
We compare DMSNet with 18 representative CNN, transformer, and pretrained
vision architectures. The complete model list and implementation details are provided in the supplementary material. All baseline models were fully fine-tuned on the target training set. DINOv3, Endo-FM, and ViT-L/16 (MAE) were evaluated using a linear probing protocol, where the pretrained backbone was frozen and only a linear classification head was trained. All models use $224\times224$ inputs and are trained with AdamW (batch size 16, learning rate $10^{-4}$, weight decay $10^{-4}$) for up to 100 epochs, with early stopping based on validation MCC. For the controlled architecture comparison, all models including DMSNet are trained using unweighted focal loss with $\gamma=2$. All experiments are repeated using three random seeds (41, 42, and 43), and results are reported as mean $\pm$ standard deviation.

\subsection{Validation strategies }
\subsubsection{Random frame-level validation}
Frames from the official Kvasir-Capsule training pool are randomly partitioned into training and validation subsets. Since the partition is performed at the frame level, visually correlated frames originating from the same source video may occur in both training and validation sets. 
\subsection{Source-Video-Disjoint Validation and Test Evaluation}
\label{sec:test_overlap}
\begin{figure*}[htbp]
    \centering
    \includegraphics[width=\textwidth]{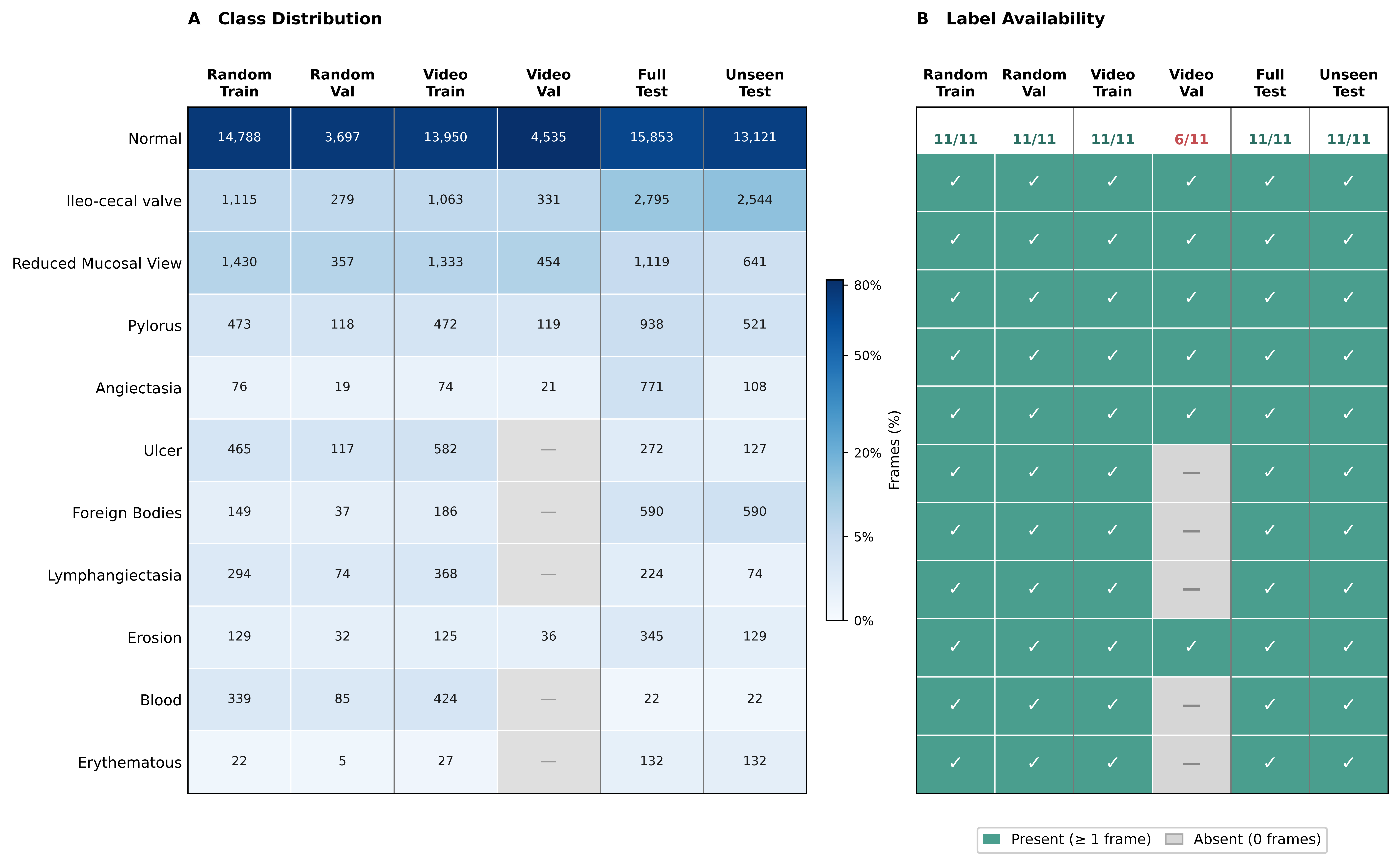}
    \caption{Class distribution and label availability
    across Kvasir-Capsule partitions.
    (A) Frame counts are annotated within each cell,
    with color representing the within-partition
    class proportion using square-root color scaling.
    (B) Class availability across random and
    source-video-disjoint training/validation
    partitions, the full official test set,
    and the source-unseen test subset.
    Source-video-disjoint validation contains
    six of the eleven classes, whereas both
    test partitions retain all eleven classes.}
    \label{fig:partition_heatmap}
    \vspace{-1.5em}
\end{figure*}

To mitigate potential data leakage arising from shared source videos, we adopt source-video-disjoint validation and test evaluation. For validation, all frames from a given source video are assigned to a single partition. Within the official Kvasir-Capsule training pool, Blood, Erythematous, Lymphangiectasia, and Ulcer each originate from a single source video, while Foreign Bodies originates from two. These rare classes are therefore retained entirely in training. The remaining multi-video classes are partitioned using four-fold \texttt{StratifiedGroupKFold}, with source-video identifiers as grouping variables, yielding 15 training and 4 validation videos with no source-video overlap. Since five rare classes are absent from validation, validation MCC is used for early stopping and checkpoint selection rather than as a comprehensive estimate of 11-class performance. Additionally, an examination of the official Kvasir-Capsule splits revealed that seven source-video identifiers are shared between the official training and test sets, accounting for approximately 22\% of the original test frames. To assess generalization without source-video overlap, we construct a source-video-disjoint test subset by excluding test frames associated with these shared identifiers. The resulting subset contains approximately 18,000 frames, with no source-video identifiers shared with the training set. All models are evaluated on both the full official test set and the source-video-disjoint test subset. Figure~\ref{fig:partition_heatmap} illustrates the class distributions and label availability across random and source-video-disjoint training/validation partitions, the full official test set, and the source-video-disjoint test subset. While random validation includes all 11 classes, source-video-disjoint validation contains only six, as recordings from five rare classes are retained in training. Both test partitions preserve all 11 classes, although excluding overlapping source videos alters their class distributions.

\subsection{Source-Video Diversity Experiment}
We vary the maximum number of source videos sampled per class and seed as $k\in\{1,2,4,8\}$. Training is capped at 200 frames per class, distributed as evenly as possible across the selected videos; when fewer frames are available, all are retained. The resulting training sets contain 1434, 1723, 2010, and 2010 frames for $k=1,2,4,$ and $8$, respectively.
Thus, lower-$k$ comparisons vary in both frame count and source-video coverage, whereas $k=4$ versus $k=8$ provides the cleanest controlled comparison at a fixed total of 2010 frames. Only 5 of 11 classes contain more than four source videos, so this comparison changes source-video
coverage only for those classes. Complete results are provided in the
supplementary material.
\subsection{Cross-Dataset Evaluation}
To assess external generalization, models trained on Kvasir-Capsule dataset using random frame-level validation are evaluated on held out test set of  CV2024 dataset. CV2024  although uses mixed-source
training data, they have seperate private held-out test set. Evaluation is restricted to the eight categories shared by the two datasets. The three Kvasir-Capsule classes that are absent in CV2024: Ileo-cecal Valve, Pylorus, and Reduced Mucosal View are masked from the prediction space before the final argmax, restricting predictions to the eight shared categories.
\subsection{Evaluation Metrics}
The performance of the proposed DMSNet and the baseline models was evaluated using four widely adopted classification metrics: Matthews Correlation Coefficient (MCC), macro recall, macro F1-score, and overall accuracy.

\section{Results}
\label{sec:results}

\subsection{Computational Efficiency}
Table~\ref{tab:efficiency} compares the computational requirements of the evaluated architectures. DMSNet contains 22.01M parameters and requires 7.561 GFLOPs per frame, while achieving 565 frames/s and 1.770 ms per-image latency on a single NVIDIA Tesla V100. Although lightweight CNNs are more efficient, DMSNet remains computationally lighter than several larger transformer-based models, supporting its practical use for large-scale VCE frame analysis.
\begin{table}[htbp]
\centering
\caption{Computational efficiency comparison at $224\times224$ input
resolution. Computational cost is estimated using THOP and reported as
GFLOPs with the convention $1$ MAC $=2$ FLOPs. Throughput and per-image
latency are measured on a single NVIDIA Tesla V100 GPU.}
\label{tab:efficiency}
\setlength{\tabcolsep}{3.5pt}
\renewcommand{\arraystretch}{1.08}

\begin{tabular}{lrrrrr}
\toprule
\textbf{Model} &
\textbf{Params (M)} &
\textbf{Trainable (M)} &
\textbf{GFLOPs} &
\textbf{Throughput} &
\textbf{Latency (ms/img)} \\
\midrule
ResNet50  &
23.53 & 23.53 & 8.26 & 1000/s & 1.000 \\

MobileNetV3-L &
4.22 & 4.22 & 0.43 & 2642/s & 0.378 \\

DenseNet-121 &
6.97 & 6.97 & 5.67 & 851/s & 1.176 \\

RegNetY-16GF &
80.60 & 80.60 & 31.82 & 265/s & 3.781 \\

EfficientNetV2-S &
20.19 & 20.19 & 5.42 & 897/s & 1.115 \\

ConvNeXtV2 &
87.70 & 87.70 & 30.71 & 234/s & 4.278 \\

ConvNeXt-Tiny &
27.83 & 27.83 & 8.91 & 797/s & 1.254 \\

ViT-B/16  &
85.81 & 85.81 & 33.70 & 310/s & 3.229 \\

MaxViT &
118.71 & 118.71 & 46.69 & 134/s & 7.461 \\

ViT-L/16 (MAE) &
303.31 & 0.01 & 119.29 &
92/s & 10.819 \\

DeiT &
85.82 & 85.82 & 33.70 & 304/s & 3.288 \\

CaiT-S24 &
46.54 & 46.54 & 17.27 & 312/s & 3.202 \\

SwinV1 &
86.75 & 86.75 & 30.34 & 249/s & 4.010 \\

SwinV2 &
48.93 & 48.93 & 17.68 & 352/s & 2.844 \\

DINOv3 &
85.65 & \textbf{0.01} & 34.37 & 278/s & 3.599 \\

Endo-FM &
121.27 & \textbf{0.01} & 47.61 & 201/s & 4.986 \\

EVA-02 &
85.77 & 85.77 & 43.89 & 191/s & 5.239 \\
\midrule

\textbf{DMSNet} &
22.01 & 22.01 & 7.561 &
565/s & 1.770 \\
\bottomrule
\end{tabular}
\end{table}

\begin{table*}[t]
\centering
\caption{Performance comparison under random and source-video-disjoint validation splits. Results are reported as mean $\pm$ standard deviation over three random seeds. Higher values indicate better performance.}
\label{tab:seen_test}
\resizebox{\textwidth}{!}{
\begin{tabular}{lcccccccc}
\toprule
{\textbf{Model}} &
\multicolumn{4}{c}{\textbf{Random Validation Split}} &
\multicolumn{4}{c}{\textbf{Source-Video-Disjoint Validation Split}} \\
\cmidrule(lr){2-5}
\cmidrule(lr){6-9}
& \textbf{MCC} & \textbf{Recall} & \textbf{F1} & \textbf{Acc.}
& \textbf{MCC} & \textbf{Recall} & \textbf{F1} & \textbf{Acc.} \\
\midrule
ResNet50
& 0.347$\pm$0.021 & 0.266$\pm$0.022 & 0.241$\pm$0.008 & 0.690$\pm$0.012
& 0.302$\pm$0.020 & 0.199$\pm$0.005 & 0.184$\pm$0.000 & 0.690$\pm$0.002 \\
MobileNetV3-L
& 0.386$\pm$0.022 & 0.258$\pm$0.011 & 0.251$\pm$0.004 & 0.711$\pm$0.004
& 0.369$\pm$0.031 & 0.246$\pm$0.023 & 0.226$\pm$0.020 & 0.700$\pm$0.011 \\
DenseNet-121
& 0.375$\pm$0.013 & 0.253$\pm$0.012 & 0.233$\pm$0.004 & 0.694$\pm$0.011
& 0.367$\pm$0.014 & 0.255$\pm$0.011 & 0.232$\pm$0.011 & 0.698$\pm$0.005 \\
RegNetY-16GF
& 0.369$\pm$0.018 & 0.289$\pm$0.012 & 0.278$\pm$0.002 & 0.700$\pm$0.010
& 0.367$\pm$0.057 & 0.272$\pm$0.023 & 0.247$\pm$0.030 & 0.691$\pm$0.027 \\
EfficientNetV2-S
& 0.398$\pm$0.028 & 0.278$\pm$0.008 & 0.278$\pm$0.010 & 0.720$\pm$0.011
& 0.388$\pm$0.016 & 0.266$\pm$0.020 & 0.265$\pm$0.030 & 0.720$\pm$0.007 \\
EfficientNet-B7
& 0.397$\pm$0.027 & 0.296$\pm$0.032 & 0.272$\pm$0.035 & 0.713$\pm$0.024
& 0.426$\pm$0.049 & 0.308$\pm$0.029 & 0.285$\pm$0.034 & 0.720$\pm$0.026 \\
ConvNeXtV2
& 0.335$\pm$0.016 & 0.240$\pm$0.025 & 0.223$\pm$0.010 & 0.694$\pm$0.008
& 0.319$\pm$0.022 & 0.225$\pm$0.014 & 0.235$\pm$0.020 & 0.713$\pm$0.006 \\
ConvNeXt-Tiny
& 0.396$\pm$0.018 & 0.280$\pm$0.022 & 0.279$\pm$0.020 & 0.721$\pm$0.007
& 0.390$\pm$0.029 & 0.266$\pm$0.027 & 0.286$\pm$0.036 & 0.729$\pm$0.013 \\
ViT-B/16
& 0.327$\pm$0.027 & 0.255$\pm$0.024 & 0.245$\pm$0.017 & 0.698$\pm$0.003
& 0.283$\pm$0.025 & 0.237$\pm$0.036 & 0.219$\pm$0.043 & 0.676$\pm$0.015 \\
MaxViT
& 0.398$\pm$0.024 & 0.278$\pm$0.010 & 0.288$\pm$0.013 & 0.728$\pm$0.005
& \textbf{0.452$\pm$0.026} & 0.312$\pm$0.014 & \textbf{0.307$\pm$0.016} & \textbf{0.730$\pm$0.023} \\

ViT-L/16 (MAE)
& 0.150$\pm$0.004 & 0.113$\pm$0.002 & 0.113$\pm$0.003 & 0.686$\pm$0.000
& 0.174$\pm$0.006 & 0.119$\pm$0.001 & 0.118$\pm$0.000 & 0.695$\pm$0.002 \\
DeiT
& 0.347$\pm$0.019 & 0.235$\pm$0.017 & 0.240$\pm$0.019 & 0.709$\pm$0.010
& 0.301$\pm$0.012 & 0.224$\pm$0.009 & 0.223$\pm$0.013 & 0.695$\pm$0.020 \\
CaiT-S24
& 0.374$\pm$0.011 & 0.269$\pm$0.037 & 0.276$\pm$0.030 & 0.724$\pm$0.006
& 0.377$\pm$0.034 & 0.250$\pm$0.031 & 0.262$\pm$0.036 & 0.728$\pm$0.011 \\
SwinV1
& 0.417$\pm$0.019 & 0.273$\pm$0.041 & 0.269$\pm$0.029 & 0.732$\pm$0.005
& 0.365$\pm$0.046 & 0.245$\pm$0.032 & 0.256$\pm$0.031 & 0.717$\pm$0.013 \\
SwinV2
& 0.378$\pm$0.036 & 0.235$\pm$0.004 & 0.255$\pm$0.009 & 0.726$\pm$0.009
& 0.372$\pm$0.041 & 0.245$\pm$0.030 & 0.248$\pm$0.032 & 0.718$\pm$0.012 \\
DINOv3
& 0.407$\pm$0.002 & 0.229$\pm$0.002 & 0.235$\pm$0.002 & 0.736$\pm$0.001
& 0.307$\pm$0.047 & 0.160$\pm$0.035 & 0.165$\pm$0.031 & 0.712$\pm$0.005 \\
Endo-FM
& 0.303$\pm$0.007 & 0.192$\pm$0.005 & 0.198$\pm$0.003 & 0.697$\pm$0.001
& 0.295$\pm$0.021 & 0.178$\pm$0.015 & 0.180$\pm$0.013 & 0.700$\pm$0.001 \\
EVA-02
& 0.315$\pm$0.008 & 0.203$\pm$0.013 & 0.199$\pm$0.010 & 0.698$\pm$0.003
& 0.313$\pm$0.019 & 0.283$\pm$0.060 & 0.239$\pm$0.049 & 0.689$\pm$0.017 \\
\midrule




\textbf{DMSNet} 
& $\mathbf{0.442 \pm 0.020}$
& $\mathbf{0.321 \pm 0.016}$
& $\mathbf{0.325 \pm 0.015}$
& $\mathbf{0.740 \pm 0.007}$
& $0.426 \pm 0.050$
& $0.299 \pm 0.024$
& $0.290 \pm 0.026$
& $0.724 \pm 0.025$
\\
\bottomrule
\end{tabular}
}
\end{table*}

\begin{table*}[!t]
\centering
\caption{Performance comparison on the source-video-disjoint test subset under random and source-video-disjoint validation splits. Results are reported as mean $\pm$ standard deviation over three random seeds. Higher values indicate better performance.}
\label{tab:unseen_test}
\resizebox{\textwidth}{!}{
\begin{tabular}{lcccccccc}
\toprule
\textbf{Model} &
\multicolumn{4}{c}{\textbf{Random Validation Split}} &
\multicolumn{4}{c}{\textbf{Source-Video-Disjoint Validation Split}} \\
\cmidrule(lr){2-5}
\cmidrule(lr){6-9}
& \textbf{MCC} & \textbf{Recall} & \textbf{F1} & \textbf{Acc.}
& \textbf{MCC} & \textbf{Recall} & \textbf{F1} & \textbf{Acc.} \\
\midrule

ResNet50
& $0.369\pm0.033$ & $0.266\pm0.034$ & $0.231\pm0.019$ & $0.723\pm0.016$
& $0.334\pm0.026$ & $0.197\pm0.004$ & $0.170\pm0.003$ & $0.729\pm0.004$ \\

MobileNetV3-L
& $0.417\pm0.021$ & $0.223\pm0.005$ & $0.202\pm0.008$ & $0.744\pm0.009$
& $0.386\pm0.030$ & $0.215\pm0.017$ & $0.186\pm0.015$ & $0.731\pm0.010$ \\

DenseNet-121
& $0.403\pm0.017$ & $0.228\pm0.009$ & $0.199\pm0.012$ & $0.723\pm0.014$
& $0.385\pm0.012$ & $0.226\pm0.010$ & $0.200\pm0.010$ & $0.733\pm0.004$ \\

RegNetY-16GF
& $0.378\pm0.009$ & $0.257\pm0.018$ & $0.232\pm0.009$ & $0.727\pm0.009$
& $0.374\pm0.051$ & $0.227\pm0.013$ & $0.202\pm0.019$ & $0.718\pm0.026$ \\

EfficientNetV2-S
& $0.419\pm0.042$ & $0.273\pm0.025$ & $0.263\pm0.031$ & $0.752\pm0.013$
& $0.397\pm0.032$ & $0.235\pm0.016$ & $0.219\pm0.026$ & $0.749\pm0.012$ \\

EfficientNet-B7
& $0.404\pm0.018$ & $0.277\pm0.030$ & $0.237\pm0.023$ & $0.740\pm0.022$
& $0.434\pm0.041$ & $0.262\pm0.023$ & $0.233\pm0.023$ & $0.749\pm0.024$ \\

ConvNeXtV2
& $0.361\pm0.014$ & $0.221\pm0.022$ & $0.202\pm0.017$ & $0.732\pm0.010$
& $0.330\pm0.014$ & $0.210\pm0.017$ & $0.204\pm0.028$ & $0.748\pm0.008$ \\

ConvNeXt-Tiny
& $0.422\pm0.026$ & $0.275\pm0.034$ & $0.256\pm0.029$ & $0.754\pm0.008$
& $0.385\pm0.011$ & $0.237\pm0.023$ & $0.223\pm0.031$ & $0.751\pm0.009$ \\

ViT-B/16
& $0.342\pm0.047$ & $0.261\pm0.030$ & $0.238\pm0.027$ & $0.733\pm0.010$
& $0.282\pm0.010$ & $0.214\pm0.027$ & $0.189\pm0.032$ & $0.702\pm0.013$ \\

MaxViT
& $0.423\pm0.029$ & $0.272\pm0.008$ & $0.262\pm0.011$ & $0.763\pm0.006$
& $\mathbf{0.456\pm0.024}$ & $0.268\pm0.008$ & $0.253\pm0.008$ & $0.755\pm0.019$ \\

ViT-L/16 (MAE)
& $0.174\pm0.006$ & $0.118\pm0.003$ & $0.119\pm0.004$ & $0.727\pm0.000$
& $0.197\pm0.007$ & $0.127\pm0.002$ & $0.125\pm0.000$ & $0.735\pm0.002$ \\

DeiT
& $0.365\pm0.028$ & $0.247\pm0.016$ & $0.236\pm0.027$ & $0.742\pm0.015$
& $0.322\pm0.020$ & $0.220\pm0.010$ & $0.214\pm0.034$ & $0.732\pm0.024$ \\

CaiT-S24
& $0.396\pm0.002$ & $0.273\pm0.026$ & $0.264\pm0.029$ & $0.758\pm0.006$
& $0.406\pm0.033$ & $0.236\pm0.031$ & $0.236\pm0.034$ & $\mathbf{0.767\pm0.005}$ \\

SwinV1
& $0.461\pm0.029$ & $0.289\pm0.024$ & $\mathbf{0.280\pm0.033}$ & $0.772\pm0.010$
& $0.372\pm0.051$ & $0.207\pm0.022$ & $0.195\pm0.016$ & $0.748\pm0.011$ \\

SwinV2
& $0.403\pm0.056$ & $0.253\pm0.037$ & $0.256\pm0.044$ & $0.761\pm0.014$
& $0.406\pm0.053$ & $0.257\pm0.033$ & $0.241\pm0.042$ & $0.756\pm0.017$ \\

DINOv3
& $\mathbf{0.487\pm0.005}$ & $0.246\pm0.003$ & $0.246\pm0.004$ & $\mathbf{0.789\pm0.001}$
& $0.378\pm0.055$ & $0.174\pm0.033$ & $0.175\pm0.029$ & $0.761\pm0.008$ \\

Endo-FM
& $0.348\pm0.010$ & $0.218\pm0.011$ & $0.214\pm0.007$ & $0.736\pm0.001$
& $0.334\pm0.022$ & $0.196\pm0.015$ & $0.187\pm0.009$ & $0.737\pm0.001$ \\

EVA-02
& $0.337\pm0.005$ & $0.218\pm0.020$ & $0.197\pm0.012$ & $0.735\pm0.005$
& $0.346\pm0.019$ & $\mathbf{0.278\pm0.045}$ & $0.240\pm0.048$ & $0.732\pm0.017$ \\

\midrule

\textbf{DMSNet}
& $0.456\pm0.037$ & $\mathbf{0.289\pm0.029}$ & $0.272\pm0.033$ & $0.768\pm0.013$
& $0.454\pm0.051$ & $0.275\pm0.016$ & $\mathbf{0.256\pm0.026}$ & $0.760\pm0.026$ \\

\bottomrule
\end{tabular}
}
\end{table*}

\begin{table*}[htbp]
\centering
\caption{Cross-dataset evaluation on the CV2024 dataset~\cite{handa2024capsule}. Results are reported as mean $\pm$ standard deviation over three random seeds.}
\label{tab:cross_dataset_cv2024}

\begin{tabular}{lcccc}
\toprule
\textbf{Model} & \textbf{MCC} & \textbf{Macro Recall} & \textbf{Macro F1} & \textbf{Accuracy} \\
\midrule
ResNet50 &
$0.133 \pm 0.025$ &
$0.189 \pm 0.018$ &
$0.161 \pm 0.029$ &
$0.570 \pm 0.046$ \\
MobileNetV3-L &
$0.1115 \pm 0.0567$ &
$0.1759 \pm 0.0320$ &
$0.1631 \pm 0.0314$ &
$0.5654 \pm 0.0129$ \\
DenseNet-121 &
$0.1768 \pm 0.0165$ &
$0.2100 \pm 0.0155$ &
$0.1866 \pm 0.0172$ &
$0.6172 \pm 0.0424$ \\
RegNetY-16GF &
$0.1209 \pm 0.0409$ &
$0.1745 \pm 0.0213$ &
$0.1482 \pm 0.0242$ &
$0.5707 \pm 0.0570$ \\
EfficientNetV2-S  &
$0.1332 \pm 0.0507$ &
$0.1886 \pm 0.0295$ &
$0.1623 \pm 0.0253$ &
$0.6042 \pm 0.1196$ \\
EfficientNet-B7 &
$0.1601 \pm 0.0204$ &
$0.2078 \pm 0.0106$ &
$0.1931 \pm 0.0039$ &
$0.5729 \pm 0.0273$ \\
ConvNeXtV2 &
$0.1908 \pm 0.0381$ &
$0.2133 \pm 0.0111$ &
$0.1911 \pm 0.0101$ &
$0.6388 \pm 0.0776$ \\
ConvNeXt-Tiny &
$\mathbf{0.2072 \pm 0.0026}$ &
$0.2185 \pm 0.0164$ &
$0.1919 \pm 0.0138$ &
$0.6506 \pm 0.0777$ \\
ViT-B/16 &
$0.0981 \pm 0.0462$ &
$0.1708 \pm 0.0202$ &
$0.1674 \pm 0.0212$ &
$0.5706 \pm 0.0244$ \\
MaxViT &
$0.1946 \pm 0.0240$ &
$0.2021 \pm 0.0130$ &
$0.1937 \pm 0.0091$ &
$0.5919 \pm 0.0962$ \\
ViT-L/16 (MAE) &
$0.0804 \pm 0.0077$ &
$0.1506 \pm 0.0045$ &
$0.1457 \pm 0.0035$ &
$0.7279 \pm 0.0029$ \\
DeiT &
$0.1526 \pm 0.0286$ &
$0.1756 \pm 0.0126$ &
$0.1742 \pm 0.0200$ &
$0.6666 \pm 0.0431$ \\
CaiT-S24 &
$0.1895 \pm 0.0384$ &
$0.2119 \pm 0.0263$ &
$0.2006 \pm 0.0133$ &
$0.6552 \pm 0.0419$ \\
SwinV1 &
$0.1687 \pm 0.0322$ &
$0.2018 \pm 0.0170$ &
$0.1961 \pm 0.0158$ &
$0.6319 \pm 0.0663$ \\
SwinV2 &
$0.1221 \pm 0.0168$ &
$0.1905 \pm 0.0083$ &
$0.1741 \pm 0.0053$ &
$0.5638 \pm 0.0395$ \\
DINOv3 &
$0.1672 \pm 0.0124$ &
$0.1919 \pm 0.0061$ &
$0.1861 \pm 0.0049$ &
$0.7151 \pm 0.0149$ \\
Endo-FM &
$0.079 \pm 0.001$ &
$0.157\pm 0.007$ &
$0.141 \pm 0.009$ &
$0.719 \pm 0.018$ \\
EVA-02  &
$0.0808 \pm 0.0102$ &
$0.1704 \pm 0.0193$ &
$0.1399 \pm 0.0166$ &
$0.5687 \pm 0.0980$ \\

\midrule
\textbf{DMSNet} &
$0.1367 \pm 0.0226$ &
$0.2066 \pm 0.0049$ &
$0.1727 \pm 0.0156$ &
$0.5789 \pm 0.0507$ \\
\bottomrule
\end{tabular}
\end{table*}

\subsection{In-Domain Benchmark and Validation Strategy}
\label{sec:indomain_results}
Table~\ref{tab:seen_test} compares DMSNet with 18 representative architectures under random frame-level
and source-video-disjoint validation. In both settings, models are evaluated on the full official Kvasir-Capsule test set. Under random frame-level validation, DMSNet achieves the highest MCC of $0.442 \pm 0.020$, followed by Swin-V1 ($0.417 \pm 0.019$) and DINOv3 ($0.407 \pm 0.002$). DMSNet also achieves a macro recall of $0.321 \pm 0.016$ and a macro F1-score of $0.325 \pm 0.015$. Under source-video-disjoint validation, MaxViT achieves the highest MCC of $0.452 \pm 0.026$, while DMSNet obtains $0.426 \pm 0.050$. Model performance varies across validation strategies, indicating that model rankings are sensitive to the validation protocol. Table~\ref{tab:unseen_test} reports the corresponding results on the source-video-disjoint test subset, constructed by excluding recordings whose source-video identifiers overlap
with the original training set. Under random frame-level validation, DMSNet achieves an MCC of $0.456 \pm 0.037$, macro recall of $0.289 \pm 0.029$, and macro F1-score of
$0.272 \pm 0.033$. DINOv3 obtains the highest MCC of $0.487 \pm 0.005$ in this setting. Under source-video-disjoint validation, MaxViT achieves the highest MCC of
$0.456 \pm 0.024$, followed closely by DMSNet at $0.454 \pm 0.051$. These results show that DMSNet remains competitive on previously unseen source
recordings, although it does not consistently outperform the baseline architectures. Taken together, Tables~\ref{tab:seen_test}
and~\ref{tab:unseen_test} reveal that model rankings are sensitive to both validation strategy and test-set composition. Although DMSNet achieves the highest MCC on the full test set under random frame-level validation, DINOv3 performs best on the source-video-disjoint subset under the same validation
protocol. Under source-video-disjoint validation, MaxViT achieves the highest MCC on both test sets, with DMSNet remaining competitive. These findings highlight the importance of evaluating architectures across different validation protocols and source-video partitions rather than drawing conclusions from a single benchmark configuration.

\subsection{Effect of Source-Video Diversity}
Table~\ref{tab:video_diversity_main} summarizes representative results, with complete results for all 19 architectures provided in the
supplementary material. Under a per-class cap of 200 training frames, the $k=4$ and $k=8$ settings provide the cleanest comparison because
both contain 2,010 frames. Increasing source-video coverage from $k=4$ to $k=8$ improves MCC for 17 of 19 architectures. MaxViT improves from
$0.245\pm0.071$ to $0.311\pm0.051$, ViT-B/16 from
$0.256\pm0.034$ to $0.297\pm0.070$, and DMSNet from
$0.201\pm0.041$ to $0.290\pm0.063$. A paired one-sided Wilcoxon
signed-rank test confirms a significant positive shift across
architectures indicating that the benefit of broader source-video coverage is not specific to a single
model family.
\begin{table}[htbp]
\centering
\caption{Effect of source-video diversity under a per-class cap of 
200 training frames. Representative architectures from different model 
families are shown. Complete results for all architectures are provided 
in the supplementary material. MCC values are reported as mean $\pm$ standard deviation over three random seeds.}

\label{tab:video_diversity_main}
\setlength{\tabcolsep}{5pt}
\begin{tabular}{lcccc}
\toprule
\textbf{Model} &
$\mathbf{k=1}$ &
$\mathbf{k=2}$ &
$\mathbf{k=4}$ &
$\mathbf{k=8}$ \\
\midrule
DMSNet      & $0.112\pm0.056$ & $0.167\pm0.074$ &
              $0.201\pm0.074$ & $0.290\pm0.063$ \\
MaxViT      & $0.163\pm0.120$ & $0.179\pm0.078$ &
              $0.245\pm0.071$ & $0.311\pm0.051$ \\
ConvNeXtV2  & $0.171\pm0.039$ & $0.201\pm0.059$ &
              $0.261\pm0.031$ & $0.276\pm0.059$ \\
ViT-B/16    & $0.142\pm0.071$ & $0.195\pm0.079$ &
              $0.256\pm0.034$ & $0.297\pm0.070$ \\
DINOv3      & $0.067\pm0.054$ & $0.050\pm0.031$ &
              $0.107\pm0.041$ & $0.172\pm0.027$ \\
\bottomrule
\end{tabular}
\end{table}
\subsection{Cross-Dataset Generalization}
Models trained on Kvasir-Capsule using random frame-level validation are evaluated without additional fine-tuning on the independent, held-out CV2024 test set, restricted to the eight classes shared between the two datasets. As shown in
Table~\ref{tab:cross_dataset_cv2024}, all architectures
exhibit limited external generalization, with the
highest MCC reaching only $0.207 \pm 0.003$.
ConvNeXt-Tiny achieves the highest MCC, followed
by MaxViT ($0.195 \pm 0.024$), ConvNeXtV2
($0.191 \pm 0.038$), and DINOv3
($0.167 \pm 0.012$). DMSNet obtains an MCC of
$0.137 \pm 0.023$, a macro recall of
$0.207 \pm 0.005$, and a macro F1-score of
$0.173 \pm 0.016$, indicating limited transfer
to the external dataset.

Although, DMSNet achieves the highest MCC on the
full Kvasir-Capsule test set under random
frame-level validation but performs substantially
worse on CV2024. The relative rankings of architectures also change across datasets, suggesting that strong in-domain performance does not necessarily indicate robust external
generalization. Although differences in acquisition
conditions and dataset characteristics may
contribute to this decline, the external evaluation
also involves a restricted eight-class prediction
space rather than the original eleven-class task.
These findings emphasize the need for external
evaluation when assessing the reliability of
VCE classification models.

\subsection{Ablation Study}
To evaluate the contribution of each component,
we conduct a leave-one-module-out ablation study
while keeping the training configuration fixed.
Table~\ref{tab:dms_ablation} presents the results.
The complete DMSNet achieves an MCC of
$0.442 \pm 0.025$, and removing any of its
components reduces mean MCC. The largest
reduction is observed when deformable lesion
refinement (DLR) is removed, decreasing MCC
to $0.401$, followed by sparse feature selection
(SFS) at $0.408$, edge-guided attention (EGA)
at $0.410$, and adaptive feature fusion (AFF)
at $0.437$. These results suggest that DLR,
SFS, and EGA make substantial contributions
to the overall classification performance,
whereas AFF provides a smaller incremental
improvement. The full architecture achieves
the highest mean MCC, supporting the
complementary roles of its components.

\begin{table}[htbp]
\centering
\caption{Component-wise ablation of DMSNet.
Results are reported as mean $\pm$ standard deviation over three random seeds.}
\label{tab:dms_ablation}
\setlength{\tabcolsep}{5pt}
\begin{tabular}{lcccc}
\toprule
\textbf{Configuration} &
\textbf{MCC} &
\textbf{Macro Recall} &
\textbf{Macro F1} &
\textbf{Accuracy} \\
\midrule
\textbf{Full DMSNet}
& $\mathbf{0.442\pm0.025}$
& $\mathbf{0.321\pm0.020}$
& $\mathbf{0.324\pm0.019}$
& $\mathbf{0.740\pm0.009}$ \\

w/o DLR
& $0.401\pm0.039$
& $0.273\pm0.005$
& $0.269\pm0.017$
& $0.728\pm0.013$ \\

w/o AFF
& $0.437\pm0.015$
& $0.297\pm0.019$
& $0.305\pm0.033$
& $0.739\pm0.012$ \\

w/o SFS
& $0.408\pm0.040$
& $0.277\pm0.037$
& $0.286\pm0.040$
& $0.735\pm0.008$ \\

w/o EGA
& $0.410\pm0.019$
& $0.273\pm0.023$
& $0.291\pm0.017$
& $0.740\pm0.001$ \\
\bottomrule
\end{tabular}
\end{table}
\subsection{Class-Wise Error Analysis}
\begin{figure}[ht]
    \centering
    \includegraphics[width=\linewidth]
    {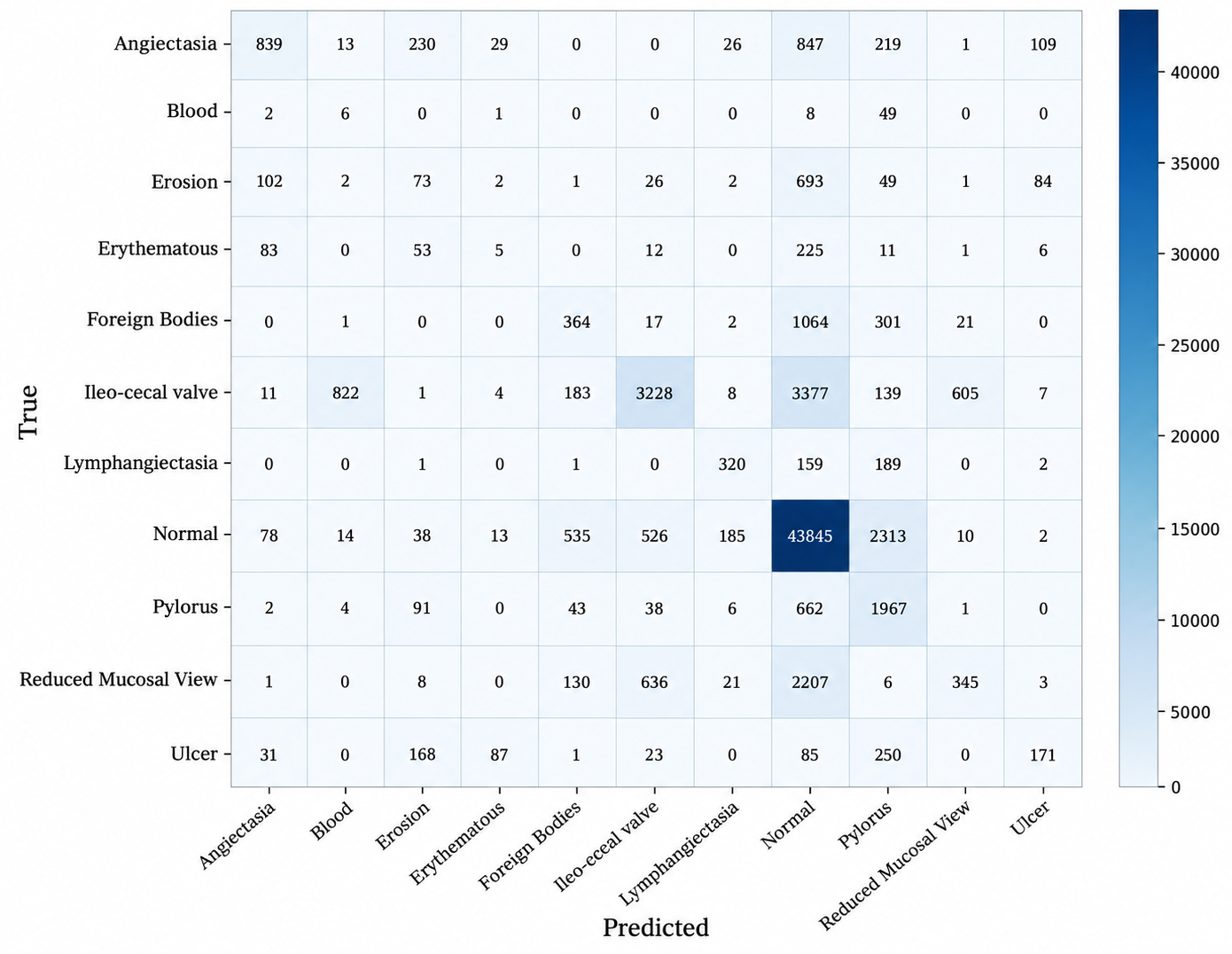}
    \caption{Confusion matrix for DMSNet.}
    \label{fig:confusion_matrix}
\end{figure}

The confusion matrix in Fig.~\ref{fig:confusion_matrix}
reveals substantial class-wise variation. DMSNet achieves
high recall for Normal (92.2\%) and Pylorus (69.9\%),
while several pathological classes are frequently confused
with normal or anatomical categories. For example,
74.2\% of Blood samples are predicted as Pylorus,
while 66.9\% of Erosion and 56.8\% of Erythematous
samples are predicted as Normal. Similarly, 60.1\%
of Foreign Bodies and 65.7\% of Reduced Mucosal View
samples are misclassified as Normal. These errors are consistent with the challenges posed by severe class imbalance and limited source-video diversity.

\subsection{Limitations and Future Work}
Several limitations remain. Source-video availability is highly imbalanced across classes, preventing all categories from contributing equally to source-video-disjoint validation and diversity experiments. The lower-$k$ diversity conditions also differ in frame count, making the fixed-size $k=4$ versus $k=8$ comparison the cleanest controlled setting. Moreover, DMSNet exhibits limited generalization to CV2024, indicating that competitive in-domain performance does not ensure robustness across independently acquired datasets. External evaluation is also restricted to the eight categories shared between Kvasir-Capsule and CV2024. Finally, our study
focuses on frame-level classification and does not model temporal information. Future work should therefore evaluate source-aware training on larger multi-center datasets and incorporate temporal modeling of complete VCE examinations.
\section{Conclusion}
This study demonstrates that VCE classification performance is sensitive to validation strategy, source-video composition,and external dataset shift. Through a systematic comparison of 19 architectures, we observe changes in model rankings across random frame-level and source-video-disjoint validation protocols. On the full Kvasir-Capsule test set, DMSNet achieves the highest MCC of $0.442 \pm 0.020$ under random frame-level validation, whereas MaxViT achieves the highest MCC of $0.452 \pm 0.026$ under source-video-disjoint validation. Evaluation on the source-video-disjoint test subset further reveals differences in model rankings, highlighting the importance of
assessing performance on previously unseen recordings. Our source-video diversity experiments show that, with the training size fixed at 2,010 frames, increasing the maximum source-video coverage from four to eight improves MCC for 17 of 19 architectures. However, external evaluation on CV2024 reveals substantial generalization limitations, with ConvNeXt-Tiny achieving the highest MCC of $0.207 \pm 0.003$, while DMSNet obtains $0.137 \pm 0.023$. Overall, these findings emphasize that competitive in-domain performance alone is insufficient to establish robust VCE classification and underscore the importance of source-aware evaluation, training-source diversity, and independent cross-dataset testing.


\bibliographystyle{plainnat}
\bibliography{references}

\newpage
\appendix

\section{Appendix}
\subsection{Baseline Architectures}
 We compare DMSNet against 18 representative architectures covering convolutional neural networks, vision transformers, and large-scale pretrained vision models. The evaluated CNN-based models include ResNet-50~\cite{he2016deep}, DenseNet-121~\cite{huang2017densely}, MobileNetV3-L~\cite{howard2019searching}, RegNetY-16GF~\cite{radosavovic2020designing}, EfficientNet-B7~\cite{tan2019efficientnet}, EfficientNetV2-S~\cite{tan2021efficientnetv2}, ConvNeXt-Tiny~\cite{liu2022convnet}, and ConvNeXtV2~\cite{woo2023convnext}. Transformer-based architectures include ViT-B/16~\cite{dosovitskiy2020image}, DeiT~\cite{touvron2021training}, CaiT-S24~\cite{touvron2021going}, Swin-V1~\cite{liu2021swin}, Swin-V2~\cite{liu2022swin}, MaxViT~\cite{tu2022maxvit}, and ViT-L/16 pretrained with MAE. We additionally evaluate pretrained models including DINOv3~\cite{simeoni2025dinov3}, EVA-02~\cite{fang2024eva}, and Endo-FM~\cite{wang2023foundation}.

\subsection{Dataset}
\begin{figure}[ht]
    \centering
    \includegraphics[width=0.6\textwidth] 
    {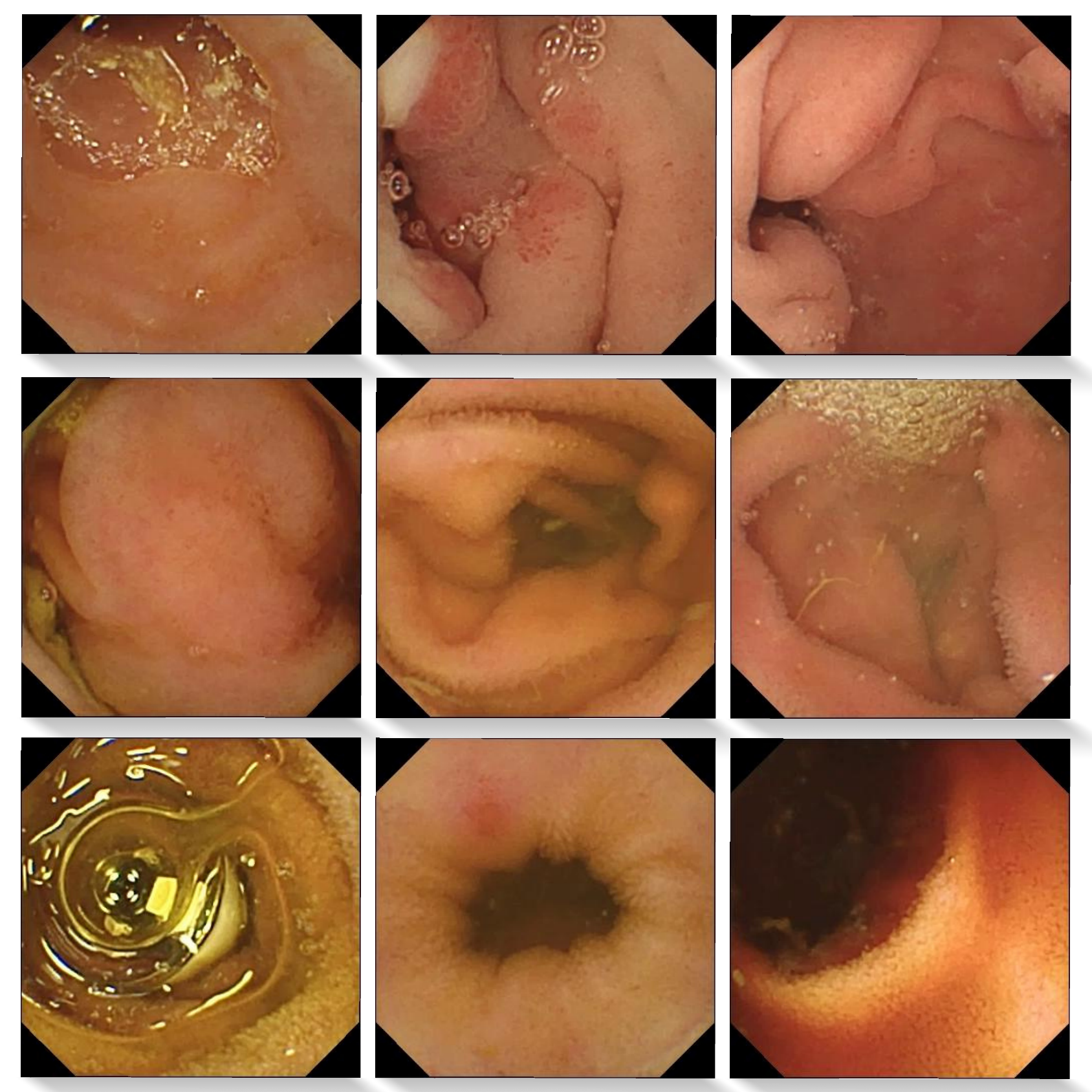}
    \caption{Representative samples from the Kvasir-Capsule dataset.}
    \label{fig:sample_images}
\end{figure}

Figure~\ref{fig:sample_images} shows representative samples from the Kvasir-Capsule dataset. 
\subsection{CV2024 label mapping}
For external evaluation, we additionally used CV2024~\cite{handa2024capsule} capsule endoscopy dataset which contains 10 classes. Kvasir-Capsule and CV2024 share eight semantically corresponding categories: Angiectasia/Angioectasia, Blood(Kvasir-Capsule Blood-fresh)/
Bleeding, Erosion, Erythematous/Erythema, Foreign Bodies/Foreign Body, Lymphangiectasia, Normal, and Ulcer. The CV2024 labels are mapped to their corresponding Kvasir-Capsule class indices before evaluation. The two CV2024 classes, Polyp and Worms, are excluded from the cross-dataset evaluation because they are not part of our 11-class Kvasir-Capsule classification setup. Polyp was excluded from the Kvasir-Capsule training task, while Worms has no corresponding Kvasir-Capsule class. Removing these two CV2024 categories excludes 38 frames in total.
\subsection{Data Augmentation}
 Since the dataset is highly imbalanced we implemented various data augmentation techniques, as increasing them using few geometric operations is essential. These operations enhanced the diversity of the datasets. We applied random horizontal flip, random vertical flip, random rotation by 15°, and color jitter with brightness 0.2 and contrast 0.2. The input images are resized to $224\times224$. 


 \subsection{Training Configuration}
 All model architectures were implemented using the PyTorch framework. The input images were resized to $(224\times224)$ pixels, and the models were trained with a batch size of 16 using the AdamW optimizer. An initial learning rate of $(1 \times 10^{-4})$ and a weight decay of $(1 \times 10^{-4})$ were employed to optimize the training process. The learning rate was dynamically adjusted using the CosineAnnealingLR scheduler throughout training.
\section{Additional Source-Video Diversity Analysis}
\subsection{Statistical Analysis of Source-Video Diversity}
\label{supp:diversity_stats}
To assess whether the improvement observed with broader source-video
coverage is consistent across architectures, we perform a paired
one-sided Wilcoxon signed-rank test on the model-level MCC values.
We focus on the $k=4$ and $k=8$ conditions because both contain exactly
2,010 training frames, making this the cleanest comparison of increased
source-video coverage without an accompanying increase in the total
training-set size. Each of the 19 evaluated architectures constitutes one
paired observation. Table~\ref{tab:paired_diversity} represents the architecture wise mcc differences between  $k=4$ and $k=8$, whereas Table~\ref{tab:video_diversity_full_1} and table~\ref{tab:video_diversity_full_2} represents complete source video diversity results under a per class cap of 200 training frames.

\begin{table*}[t]
\centering
\caption{Complete source-video diversity results under a per-class cap of
200 training frames (Part I). Results are reported as mean $\pm$ standard
deviation over three random seeds. Bold values indicate the best result for
each metric within each architecture across the four source-video settings.}
\label{tab:video_diversity_full_1}
\small
\setlength{\tabcolsep}{5pt}

\begin{tabular}{lccccc}
\toprule
\textbf{Model} &
\textbf{\# Videos} &
\textbf{MCC} &
\textbf{Macro Recall} &
\textbf{Macro F1} &
\textbf{Accuracy} \\
\midrule




\multirow{4}{*}{ResNet50}
& 1 & $0.076 \pm 0.053$ & $0.165 \pm 0.038$ & $0.082 \pm 0.038$ & $0.187 \pm 0.033$ \\
& 2 & $0.131 \pm 0.111$ & $0.161 \pm 0.054$ & $0.131 \pm 0.053$ & $0.275 \pm 0.136$ \\
& 4 & $0.195 \pm 0.035$ & $0.156 \pm 0.016$ & $0.149 \pm 0.012$ & $0.387 \pm 0.038$ \\
& 8 & $\mathbf{0.221 \pm 0.049}$ & $\mathbf{0.189 \pm 0.031}$ & $\mathbf{0.158 \pm 0.024}$ & $\mathbf{0.432 \pm 0.064}$ \\
\midrule

\multirow{4}{*}{EfficientNetV2-S}
& 1 & $0.076 \pm 0.046$ & $0.188 \pm 0.015$ & $0.099 \pm 0.021$ & $0.159 \pm 0.059$ \\
& 2 & $0.145 \pm 0.060$ & $0.192 \pm 0.022$ & $0.140 \pm 0.033$ & $0.283 \pm 0.105$ \\
& 4 & $0.194 \pm 0.049$ & $0.184 \pm 0.029$ & $0.152 \pm 0.016$ & $0.412 \pm 0.054$ \\
& 8 & $\mathbf{0.239 \pm 0.030}$ & $\mathbf{0.240 \pm 0.007}$ & $\mathbf{0.189 \pm 0.014}$ & $\mathbf{0.470 \pm 0.042}$ \\
\midrule

\multirow{4}{*}{ConvNeXtV2}
& 1 & $0.171 \pm 0.039$ & $\mathbf{0.261 \pm 0.014}$ & $0.160 \pm 0.023$ & $0.335 \pm 0.069$ \\
& 2 & $0.201 \pm 0.059$ & $0.233 \pm 0.014$ & $0.189 \pm 0.018$ & $0.362 \pm 0.160$ \\
& 4 & $0.261 \pm 0.031$ & $0.254 \pm 0.043$ & $0.202 \pm 0.015$ & $0.477 \pm 0.037$ \\
& 8 & $\mathbf{0.276 \pm 0.059}$ & $0.250 \pm 0.033$ & $\mathbf{0.217 \pm 0.046}$ & $\mathbf{0.509 \pm 0.105}$ \\
\midrule
\multirow{4}{*}{MobileNetV3-L}
& 1 & $0.084 \pm 0.053$ & $0.203 \pm 0.029$ & $0.116 \pm 0.028$ & $0.231 \pm 0.053$ \\
& 2 & $0.175 \pm 0.079$ & $0.192 \pm 0.055$ & $0.150 \pm 0.036$ & $0.341 \pm 0.140$ \\
& 4 & $0.195 \pm 0.060$ & $0.161 \pm 0.016$ & $0.153 \pm 0.014$ & $0.406 \pm 0.073$ \\
& 8 & $\mathbf{0.282 \pm 0.059}$ & $\mathbf{0.216 \pm 0.016}$ & $\mathbf{0.195 \pm 0.018}$ & $\mathbf{0.525 \pm 0.079}$ \\
\midrule

\multirow{4}{*}{DenseNet-121}
& 1 & $0.107 \pm 0.059$ & $\mathbf{0.239 \pm 0.042}$ & $0.122 \pm 0.051$ & $0.199 \pm 0.055$ \\
& 2 & $0.192 \pm 0.069$ & $0.232 \pm 0.025$ & $0.183 \pm 0.035$ & $0.301 \pm 0.140$ \\
& 4 & $0.200 \pm 0.079$ & $0.207 \pm 0.042$ & $0.178 \pm 0.038$ & $0.377 \pm 0.074$ \\
& 8 & $\mathbf{0.244 \pm 0.053}$ & $0.218 \pm 0.022$ & $\mathbf{0.190 \pm 0.024}$ & $\mathbf{0.468 \pm 0.073}$ \\
\midrule

\multirow{4}{*}{RegNetY-16GF}
& 1 & $0.129 \pm 0.044$ & $0.261 \pm 0.050$ & $0.144 \pm 0.027$ & $0.236 \pm 0.038$ \\
& 2 & $0.196 \pm 0.062$ & $0.221 \pm 0.013$ & $0.158 \pm 0.026$ & $0.352 \pm 0.169$ \\
& 4 & $0.192 \pm 0.063$ & $0.208 \pm 0.052$ & $0.168 \pm 0.036$ & $0.417 \pm 0.036$ \\
& 8 & $\mathbf{0.250 \pm 0.023}$ & $\mathbf{0.266 \pm 0.039}$ & $\mathbf{0.179 \pm 0.016}$ & $\mathbf{0.501 \pm 0.033}$ \\
\midrule

\multirow{4}{*}{EfficientNet-B7}
& 1 & $0.141 \pm 0.044$ & $0.211 \pm 0.016$ & $0.128 \pm 0.020$ & $0.305 \pm 0.013$ \\
& 2 & $0.192 \pm 0.034$ & $0.227 \pm 0.019$ & $0.169 \pm 0.013$ & $0.355 \pm 0.086$ \\
& 4 & $0.196 \pm 0.055$ & $0.232 \pm 0.027$ & $0.168 \pm 0.014$ & $0.414 \pm 0.067$ \\
& 8 & $\mathbf{0.276 \pm 0.069}$ & $\mathbf{0.256 \pm 0.027}$ & $\mathbf{0.212 \pm 0.028}$ & $\mathbf{0.494 \pm 0.077}$ \\
\midrule

\multirow{4}{*}{ConvNeXt-Tiny}
& 1 & $0.185 \pm 0.041$ & $\mathbf{0.263 \pm 0.012}$ & $0.154 \pm 0.026$ & $0.363 \pm 0.047$ \\
& 2 & $0.192 \pm 0.048$ & $0.247 \pm 0.023$ & $0.181 \pm 0.024$ & $0.344 \pm 0.076$ \\
& 4 & $0.249 \pm 0.067$ & $0.203 \pm 0.028$ & $0.181 \pm 0.015$ & $0.456 \pm 0.100$ \\
& 8 & $\mathbf{0.284 \pm 0.099}$ & $0.226 \pm 0.046$ & $\mathbf{0.205 \pm 0.034}$ & $\mathbf{0.508 \pm 0.122}$ \\
\midrule

\multirow{4}{*}{ViT-B/16}
& 1 & $0.142 \pm 0.071$ & $0.225 \pm 0.068$ & $0.128 \pm 0.022$ & $0.303 \pm 0.063$ \\
& 2 & $0.195 \pm 0.079$ & $0.229 \pm 0.055$ & $0.184 \pm 0.044$ & $0.324 \pm 0.150$ \\
& 4 & $0.256 \pm 0.034$ & $0.209 \pm 0.036$ & $0.187 \pm 0.010$ & $0.449 \pm 0.013$ \\
& 8 & $\mathbf{0.297 \pm 0.070}$ & $\mathbf{0.279 \pm 0.055}$ & $\mathbf{0.234 \pm 0.042}$ & $\mathbf{0.540 \pm 0.041}$ \\
\midrule

\multirow{4}{*}{DeiT}
& 1 & $0.150 \pm 0.041$ & $0.239 \pm 0.030$ & $0.135 \pm 0.026$ & $0.339 \pm 0.120$ \\
& 2 & $0.210 \pm 0.049$ & $0.220 \pm 0.004$ & $0.178 \pm 0.021$ & $0.356 \pm 0.107$ \\
& 4 & $\mathbf{0.252 \pm 0.007}$ & $0.235 \pm 0.012$ & $0.196 \pm 0.017$ & $\mathbf{0.494 \pm 0.045}$ \\
& 8 & $0.251 \pm 0.030$ & $\mathbf{0.313 \pm 0.039}$ & $\mathbf{0.228 \pm 0.029}$ & $0.485 \pm 0.054$ \\
\midrule

\multirow{4}{*}{SwinV2}
& 1 & $0.134 \pm 0.040$ & $0.243 \pm 0.010$ & $0.132 \pm 0.015$ & $0.210 \pm 0.047$ \\
& 2 & $0.278 \pm 0.099$ & $0.242 \pm 0.021$ & $0.203 \pm 0.055$ & $0.464 \pm 0.174$ \\
& 4 & $\mathbf{0.287 \pm 0.082}$ & $\mathbf{0.248 \pm 0.041}$ & $0.223 \pm 0.027$ & $0.535 \pm 0.085$ \\
& 8 & $0.285 \pm 0.054$ & $0.242 \pm 0.027$ & $\mathbf{0.226 \pm 0.035}$ & $\mathbf{0.578 \pm 0.057}$ \\
\midrule

\multirow{4}{*}{ViT-L/16 (MAE)}
& 1 & $0.090 \pm 0.039$ & 
$0.147 \pm 0.016$ & 
$0.082 \pm 0.016$ & 
$0.277 \pm 0.121$ \\

& 2 & $0.125 \pm 0.066$
& $0.188 \pm 0.021$ 
& $0.082 \pm 0.024$ 
& $0.271 \pm 0.108$ \\

& 4 & $0.121 \pm 0.051$ & 
$0.166 \pm 0.02$ &
$0.100 \pm 0.016$ & 
$0.214 \pm 0.075$ \\

& 8 & $0.141 \pm 0.049$ & 
$0.182 \pm 0.037$ & 
$0.096 \pm 0.021$ & 
$0.206 \pm 0.080$ \\

\bottomrule
\end{tabular}
\end{table*}

\begin{table*}[!ht]
\centering
\caption{Complete source-video diversity results under a per-class cap of
200 training frames (Part II). Results are reported as mean $\pm$ standard
deviation over three random seeds. Bold values indicate the best result for
each metric within each architecture across the four source-video settings.}
\label{tab:video_diversity_full_2}
\small
\setlength{\tabcolsep}{5pt}

\begin{tabular}{lccccc}
\toprule
\textbf{Model} &
\textbf{\# Videos} &
\textbf{MCC} &
\textbf{Macro Recall} &
\textbf{Macro F1} &
\textbf{Accuracy} \\
\midrule
\multirow{4}{*}{DMSNet}
& 1 & $0.112 \pm 0.056$ & $\mathbf{0.199 \pm 0.011}$ & $0.123 \pm 0.018$ & $0.310 \pm 0.073$ \\
& 2 & $0.167 \pm 0.074$ & $0.217 \pm 0.057$ & $0.158 \pm 0.032$ & $0.337 \pm 0.153$ \\
& 4 & $0.201 \pm 0.077$ & $0.229 \pm 0.039$ & $0.171 \pm 0.022$ & $0.392 \pm 0.099$ \\
& 8 & $\mathbf{0.290 \pm 0.063}$ & $0.241 \pm 0.01$ & $\mathbf{0.196 \pm 0.019}$ & $\mathbf{0.564 \pm 0.049}$ \\
\midrule

\multirow{4}{*}{MaxViT}
& 1 & $0.163 \pm 0.120$ & $0.249 \pm 0.028$ & $0.155 \pm 0.053$ & $0.331 \pm 0.167$ \\
& 2 & $0.179 \pm 0.078$ & $0.227 \pm 0.052$ & $0.161 \pm 0.031$ & $0.341 \pm 0.147$ \\
& 4 & $0.245 \pm 0.071$ & $0.256 \pm 0.003$ & $0.198 \pm 0.018$ & $0.458 \pm 0.104$ \\
& 8 & $\mathbf{0.311 \pm 0.051}$ & $\mathbf{0.280 \pm 0.047}$ & $\mathbf{0.236 \pm 0.024}$ & $\mathbf{0.550 \pm 0.062}$ \\
\midrule



\multirow{4}{*}{DINOv3}
& 1 & $0.067 \pm 0.054$ & $0.165 \pm 0.055$ & $0.083 \pm 0.036$ & $0.184 \pm 0.040$ \\
& 2 & $0.050 \pm 0.031$ & $0.146 \pm 0.013$ & $0.071 \pm 0.020$ & $0.163 \pm 0.080$ \\
& 4 & $0.107 \pm 0.041$ & $0.169 \pm 0.017$ & $0.098 \pm 0.023$ & $0.222 \pm 0.061$ \\
& 8 & $\mathbf{0.172 \pm 0.027}$ & $\mathbf{0.216 \pm 0.013}$ & $\mathbf{0.129 \pm 0.014}$ & $\mathbf{0.301 \pm 0.046}$ \\
\midrule

\multirow{4}{*}{EVA-02}
& 1 & $0.174 \pm 0.038$ & $\mathbf{0.269 \pm 0.054}$ & $0.166 \pm 0.015$ & $0.340 \pm 0.065$ \\
& 2 & $0.226 \pm 0.037$ & $0.228 \pm 0.023$ & $0.183 \pm 0.025$ & $0.437 \pm 0.100$ \\
& 4 & $0.233 \pm 0.022$ & $0.198 \pm 0.045$ & $0.179 \pm 0.028$ & $\mathbf{0.482 \pm 0.026}$ \\
& 8 & $\mathbf{0.259 \pm 0.067}$ & $0.234 \pm 0.035$ & $\mathbf{0.202 \pm 0.029}$ & $0.475 \pm 0.129$ \\
\midrule

\multirow{4}{*}{Endo-FM}
& 1 & $0.067 \pm 0.051$ & $\mathbf{0.159 \pm 0.057}$ & $0.089 \pm 0.019$ & $0.203 \pm 0.021$ \\
& 2 & $0.152 \pm 0.057$ & $0.157 \pm 0.038$ & $0.139 \pm 0.024$ & $0.333 \pm 0.090$ \\
& 4 & $0.140 \pm 0.055$ & $0.140 \pm 0.024$ & $0.137 \pm 0.012$ & $0.304 \pm 0.068$ \\
& 8 & $\mathbf{0.193 \pm 0.045}$ & $0.149 \pm 0.012$ & $\mathbf{0.155 \pm 0.015}$ & $\mathbf{0.392 \pm 0.051}$ \\ 
\midrule
\multirow{4}{*}{CaiT-S24}
& 1 & $0.205 \pm 0.068$ & $0.227 \pm 0.034$ & $0.148 \pm 0.024$ & $0.396 \pm 0.072$ \\
& 2 & $0.220 \pm 0.060$ & $0.229 \pm 0.035$ & $0.202 \pm 0.056$ & $0.430 \pm 0.110$ \\
& 4 & $0.220 \pm 0.102$ & $0.218 \pm 0.058$ & $0.189 \pm 0.051$ & $0.445 \pm 0.080$ \\
& 8 & $\mathbf{0.349 \pm 0.039}$ & $\mathbf{0.263 \pm 0.021}$ & $\mathbf{0.228 \pm 0.008}$ & $\mathbf{0.604 \pm 0.024}$ \\
\midrule

\multirow{4}{*}{SwinV1}
& 1 & $0.121 \pm 0.099$ & $0.268 \pm 0.008$ & $0.139 \pm 0.038$ & $0.275 \pm 0.094$ \\
& 2 & $0.200 \pm 0.093$ & $0.232 \pm 0.051$ & $0.181 \pm 0.032$ & $0.422 \pm 0.091$ \\
& 4 & $0.271 \pm 0.032$ & $0.214 \pm 0.017$ & $0.195 \pm 0.007$ & $0.556 \pm 0.106$ \\
& 8 & $\mathbf{0.307 \pm 0.019}$ & $\mathbf{0.292 \pm 0.018}$ & $\mathbf{0.238 \pm 0.011}$ & $\mathbf{0.558 \pm 0.035}$ \\
\bottomrule
\end{tabular}
\end{table*}

\begin{table}[t]
\centering
\caption{Architecture-wise MCC differences between $k=4$ and $k=8$.
Both conditions contain 2,010 training frames. Positive values indicate
higher mean MCC with broader source-video coverage.}
\label{tab:paired_diversity}
\small
\begin{tabular}{lccc}
\toprule
\textbf{Architecture} &
\textbf{MCC ($k=4$)} &
\textbf{MCC ($k=8$)} &
\textbf{$\Delta$ MCC} \\
\midrule
DMSNet             & 0.290 & 0.201 & $+0.089$ \\
MaxViT             & 0.245 & 0.311 & $+0.066$ \\
DINOv3             & 0.107 & 0.172 & $+0.065$ \\
ResNet50           & 0.195 & 0.221 & $+0.026$ \\
EfficientNetV2-S   & 0.194 & 0.239 & $+0.045$ \\
ConvNeXtV2         & 0.261 & 0.276 & $+0.015$ \\
ViT-B/16           & 0.256 & 0.297 & $+0.041$ \\
DeiT               & 0.252 & 0.251 & $-0.001$ \\
SwinV2             & 0.287 & 0.285 & $-0.002$ \\
EVA-02             & 0.233 & 0.259 & $+0.026$ \\
Endo-FM            & 0.140 & 0.193 & $+0.053$ \\
MobileNetV3-L      & 0.195 & 0.282 & $+0.087$ \\
DenseNet121        & 0.200 & 0.244 & $+0.044$ \\
RegNetY-16GF       & 0.192 & 0.250 & $+0.058$ \\
EfficientNet-B7    & 0.196 & 0.276 & $+0.080$ \\
ConvNeXt-Tiny      & 0.249 & 0.284 & $+0.035$ \\
CaiT-S24           & 0.220 & 0.349 & $+0.129$ \\
SwinV1             & 0.271 & 0.307 & $+0.036$ \\
ViT-L/16 (MAE)     & 0.121 & 0.141 & $+0.02$\\

\bottomrule
\end{tabular}
\end{table}



\end{document}